\documentclass[letterpaper, 10 pt, conference]{ieeeconf}  

\IEEEoverridecommandlockouts                              
                                                          
\usepackage{xcolor}
\usepackage{multirow}

\usepackage{color}
\usepackage{booktabs}
\usepackage{array}

\usepackage{amssymb}
\usepackage{amsmath}
\usepackage{amstext}

\usepackage{multicol}
\usepackage{multirow}
\usepackage{colortbl}
\usepackage{xcolor}
\usepackage{booktabs}   
\usepackage{bbding} 

\newcommand{\ourcell}{\cellcolor{gray!10}}
\newcommand{\ci}[1]{\tiny{\textcolor{gray}{~($\pm #1$)}}}

\usepackage{arydshln}

\usepackage{adjustbox}
\usepackage[font=footnotesize,labelfont=bf]{caption}

\definecolor{citecolor}{HTML}{3953A4} 
\definecolor{lblue}{HTML}{0071bc} 
\definecolor{ogreen}{HTML}{2E7D32}
\definecolor{bred}{HTML}{BF360C}
\definecolor{newbrown}{HTML}{795548}

\newcolumntype{C}[1]{>{\centering\arraybackslash}p{#1}}

\def\ourmethod{{DEX}}

\def\ebuff{{\mathcal{D}_E}}
\def\abuff{{\mathcal{D}_A}}
\def\ast{{s_t}}

\def\epolicy{{\pi^e}}
\def\eaction{a^e}

\usepackage{hyperref}
\makeatletter
\let\NAT@parse\undefined
\makeatother
\usepackage[numbers,sort,compress]{natbib}
\usepackage[all=normal,bibbreaks=tight,paragraphs=tight,floats=tight]{savetrees}
\hypersetup{
colorlinks=true, linkcolor=bred, urlcolor=lblue, citecolor=citecolor,
}

\title{\LARGE \bf 
Demonstration-Guided Reinforcement Learning with \\ Efficient Exploration for Task Automation of Surgical Robot 
}

\author{Tao Huang$^{1}$, Kai Chen$^{1}$, Bin Li$^{2}$, Yun-Hui Liu$^{2}$, Qi Dou$^{1}$
\thanks{
This research work was supported in part by CUHK T Stone Robotics Institute, Hong Kong Innovation and Technology Commission Project No. ITS/237/21FP, Hong Kong Research Grants Council TRS Project No.T42-409/18-R, and InnoHK Multi-Scale Medical Robotics Center.}
\thanks{$^{1}$T. Huang, K. Chen, and Q. Dou are with the Department of Computer Science and Engineering, The Chinese University of Hong Kong.}
\thanks{$^{2}$B. Li, and Y. H. Liu are with the Department of Mechanical and Automation Engineering, The Chinese University of Hong Kong.}
\thanks{Corresponding author: Qi Dou (qidou@cuhk.edu.hk).}%
}

\begin{document}

\maketitle
\thispagestyle{empty}
\pagestyle{empty}

\begin{abstract}
Task automation of surgical robot has the potentials to improve surgical efficiency. Recent reinforcement learning (RL) based approaches provide scalable solutions to surgical automation, but typically require extensive data collection to solve a
task if no prior knowledge is given. This issue is known as the exploration challenge, which can be alleviated by providing expert demonstrations to an RL agent. Yet, how to make effective use of demonstration data to improve exploration efficiency still remains an open challenge.  
In this work, we introduce Demonstration-guided EXploration (DEX), an efficient reinforcement learning algorithm that aims to overcome the exploration problem with expert demonstrations for surgical automation. To effectively exploit demonstrations, our method estimates expert-like behaviors with higher values to facilitate productive interactions, and adopts non-parametric regression to enable such guidance at states unobserved in demonstration data. Extensive experiments on $10$ surgical manipulation tasks from SurRoL, a comprehensive surgical simulation platform, demonstrate significant improvements in the exploration efficiency and task success rates of our method. Moreover, we also deploy the learned policies to the da Vinci Research Kit (dVRK) platform to show the effectiveness on the real robot. Code is available at \url{https://github.com/med-air/DEX}.

\end{abstract}

\section{Introduction}
Surgical robots nowadays assist surgeons to conduct minimally invasive interventions in clinical routine~\cite{d2021accelerating}. Automating surgical tasks has been increasingly desired to improve surgical efficiency, with many efforts being made on different tasks such as suturing~\cite{schwaner2021autonomous,srl_multistage,suture}, endoscope control~\cite{gao2022savanet,endoscope,pandya2014review}, tissue manipulation~\cite{saeidi2022autonomous,flexml,srl_reward} and pattern cutting~\cite{cutting1,nguyen2019new,nguyen2019manipulating}. Recently, reinforcement learning (RL) approaches have exhibited high scalability to learn diverse control policies and yielded promising performance in surgical automation~\cite{su2020reinforcement,srl_reward1,scheikl2021cooperative,segato2021inverse,tan2019robot,keller2020optical,d2022learning}, but typically require extensive data collection to solve a task if no prior knowledge is given. This issue, known as the exploration challenge, gives rise to the idea of providing expert knowledge from demonstration data to an RL agent~\cite{ramirez2022model}. Yet, how to make full effective of demonstrations to improve exploration efficiency still remains an open challenge.

One classical approach is to give higher sampling priority to demonstrated data over self-collected data in a simulator~\cite{levine2013guided,dqfd,ddpgfd}. While straightforward, the expert knowledge from demonstrations is not explicitly exploited during the exploration, making these methods still inefficient in robotic tasks. To provide explicit exploration guidance, one other approach offers RL agent with an additional reward function learned from demonstrations, which encourages the agent to stay close to expert samples~\cite{zhu,deepmimic,amp} by discerning expert-like and -unlike behaviors. However, these methods are not easily applicable to surgical tasks because they not only require careful task-specific design of environment reward function~\cite{tissue_lfd}, but also introduces the risk of local optima during policy learning. 

\begin{figure}[t]
    \centering
    \vspace{0.3cm}
    \centerline{\includegraphics[width=1\linewidth]{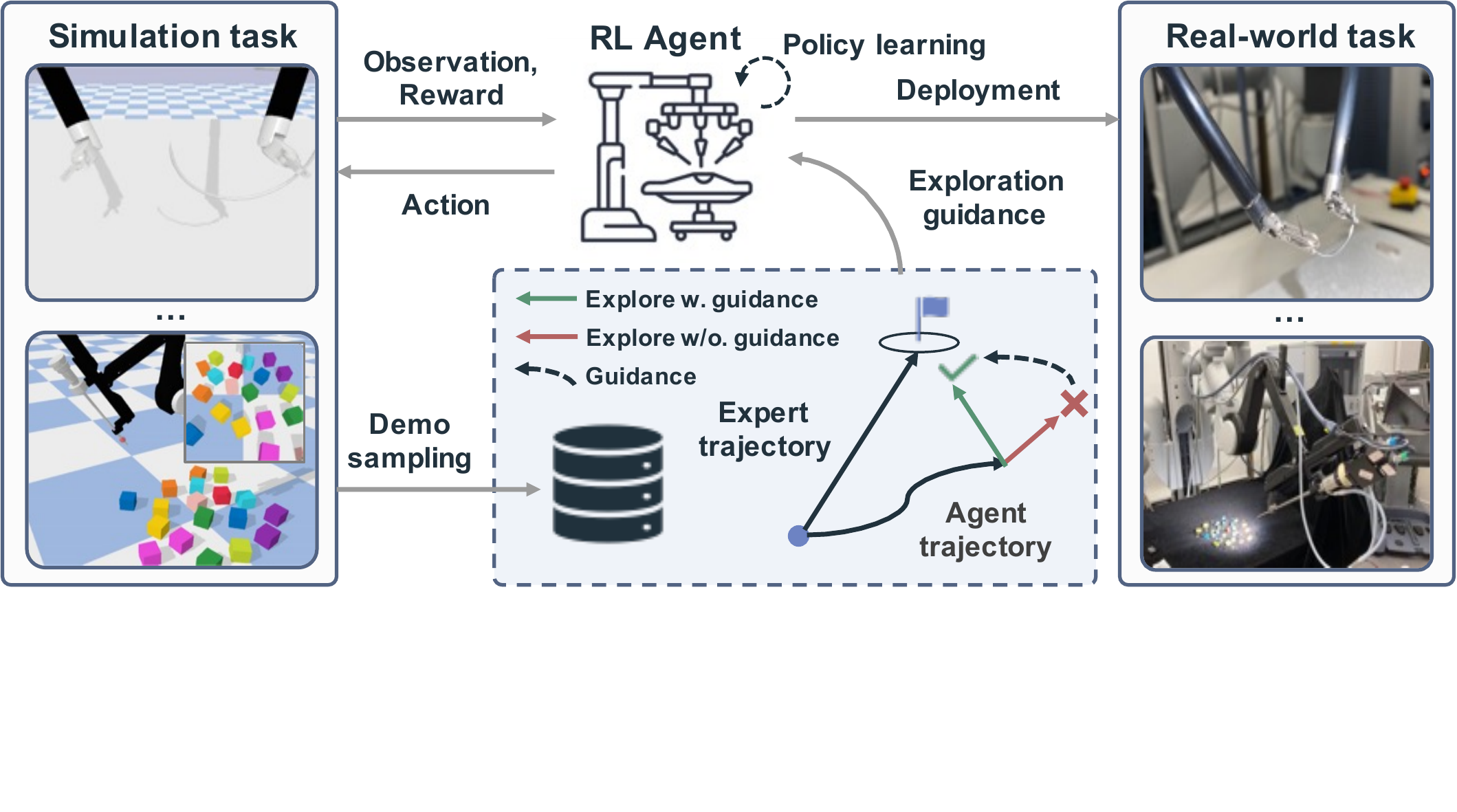}}
    \vspace{-1.35cm}
    \caption{Our method DEX leverages demonstration data to efficiently guide the RL exploration in simulation tasks. The learned policies are successfully deployed to the real-world surgery subtasks on the dVRK platform.}
    \label{fig:overview}
    \vspace{0cm}
\end{figure}

To overcome such limitations, a promising approach is to directly regularize the robot policy to mimic the expert policy within the actor-critic framework~\cite{ddpgher,dapg,zhu2019dexterous,shah2021rrl,col}. Such guidance is typically operationalized by regularizing the actor loss with a penalty measuring the behavioral dissimilarity between the robot and the expert. Effective as these methods are, their exploration efficiency is still unsatisfactory due to two limitations. Firstly, these methods enforce little regularization on the critic, making the critic suffers from the notorious overestimation issue~\cite{td3,doubleql}. Without an accurate critic, the overestimated values of expert-unlike actions hinder the robot from exploring expert-like actions. Secondly, these methods provide effective guidance only when the current state is close to the ones in the demonstrations. This impedes their capability to guide exploration early in the learning process, as the robot with a mediocre policy is likely to visit states far from demonstrated ones. A stopgap is to cover as many possible demonstrated states as possible by collecting a large number of demonstrations~\cite{chiu2021bimanual}. Unfortunately, collecting many surgical expert data is often unaffordable due to the resources and ethical constraints.

In this work, we aim to improve the exploration efficiency of RL given a modest set of demonstration data for task automation of surgical robot. To tackle the overestimation issue of the critic, we propose a novel algorithmic framework, named expert-guided actor-critic, that regularizes both actor and critic with the action dissimilarity between agent and expert. As illustrated in Figure~\ref{fig:overview}, at a high level, with this regularization the critic lowers the value estimates of expert-unlike actions (colored with red), thus further encouraging the exploration on expert-like actions (colored with green). To guide exploration at states unobserved in demonstrations, we adopt a non-parametric regression model to robustly propagate the guidance (dashed arrow) from demonstration data to these states. Consequently, our method, named Demonstration-guided EXploration (\ourmethod), is of high exploration efficiency with a limited amount of demonstration data. The empirical results on surgical automation tasks from SurRoL~\cite{surrol} show that~\ourmethod~significantly outperforms prior learning-based methods. We also validate the transferability of learned policies on da Vinci surgical robots. The contributions of this work are summarized as follows:
\begin{itemize}
    \item We propose a novel actor-critic framework to mitigate the overestimation issue of the critic for encouraging explorations on expert-like actions in RL;
    \item We adopt non-parametric guidance propagation to guide exploration at demonstration-unobserved states;
    \item We empirically demonstrate that our method significantly outperforms prior RL-based approaches on the surgical robot learning tasks from SurRoL;
    \item We successfully deploy our trained policies on the real dVRK, indicating its great potential for task automation of surgical robot in the real world.
\end{itemize}

\section{Related Work}\label{sec:related_work}
\noindent\textbf{Surgical Robot Task Automation.} Considerable efforts have been made by researchers to design skill-oriented controllers for surgical task automation, including surgical suturing~\cite{leonard2014smart,schwaner2021autonomous}, endoscope control~\cite{endoscope,gao2022savanet}, tissue manipulation~\cite{attanasio2020autonomous,saeidi2022autonomous} and pattern cutting~\cite{patel2018using,cutting2}. However, these works follow a traditional paradigm that requires domain knowledge for designing control strategies and suffers from low generality, i.e., a controller designed for one task is often hard to generalize~\cite{ras}. In contrast, learning-based approaches, representatively RL, enable robots to flexibly learn to perform tasks from collected data. These methods thus do not require task-specific control strategies and have shown improved generalization capabilities in automating complex surgical tasks~\cite{su2020reinforcement,srl_reward1,scheikl2021cooperative,segato2021inverse,tan2019robot,keller2020optical,d2022learning}. Promising as it is, running existing RL methods in surgical tasks is still inefficient due to the exploration challenge, which is often alleviated through substantial reward engineering for each task. To this end, we propose a novel RL algorithm to solve diverse tasks efficiently by exploiting expert demonstrations for facilitating task automation in robotic surgery. 

\noindent\textbf{Learning from Demonstrations.} Imitation learning (IL), also known as learning from demonstrations~\cite{argall2009survey}, is a typical learning-based framework that leverages demonstrations for policy learning. A well-known approach is behavior cloning~(BC,~\cite{bc}) which supervises the agent to imitate expert behavior. However, it suffers from the distribution shift problem and thus is of poor generalization performance~\cite{bc_compouding,osa2018algorithmic}. Some IL methods overcome this issue by using generative adversarial networks to match the trajectory distribution of demonstrations~\cite{gail, ding2019goal} or provide the agent with inferred reward function~\cite{meirl, airl}. But they require a great many online samples and demonstrations to ensure fair generalization capability and suffer from poor training stability~\cite{dac}. In contrast, our method goes beyond such limitations by utilizing the expert knowledge from demonstration data to expedite RL exploration, which combines advances of IL and RL and alleviates their respective drawbacks. 

\noindent\textbf{Demonstration-Guided RL.}
A number of demonstration-guided RL approaches have been proposed to mitigate the exploration challenges of RL, including prioritizing expert samples over self-collected samples~\cite{levine2013guided,dqfd,ddpgfd} or learning extra reward functions~\cite{zhu,deepmimic,amp}. However, these methods either make no explicit use of demonstrations or still need careful reward engineering in surgical tasks. To overcome these issues, a promising approach is to leverage demonstrations to directly regularize the RL policy to explore expert-like behaviors~\cite{ddpgher,dapg,zhu2019dexterous,shah2021rrl,col}. It penalizes the behavioral dissimilarity between agent and expert with state-action pairs from demonstrations. Nonetheless, this approach often requires a large number of demonstrations to ensure high exploration efficiency due to the impediment of overestimated values and the incapability of offering guidance at states far from the demonstrated ones. Our method addresses these limitations through critic regularization in the proposed actor-critic framework and robust guidance propagation based on non-parametric regression.

\section{Method}
\begin{figure*}[t]
\centering
\centerline{\includegraphics[width=1\textwidth]{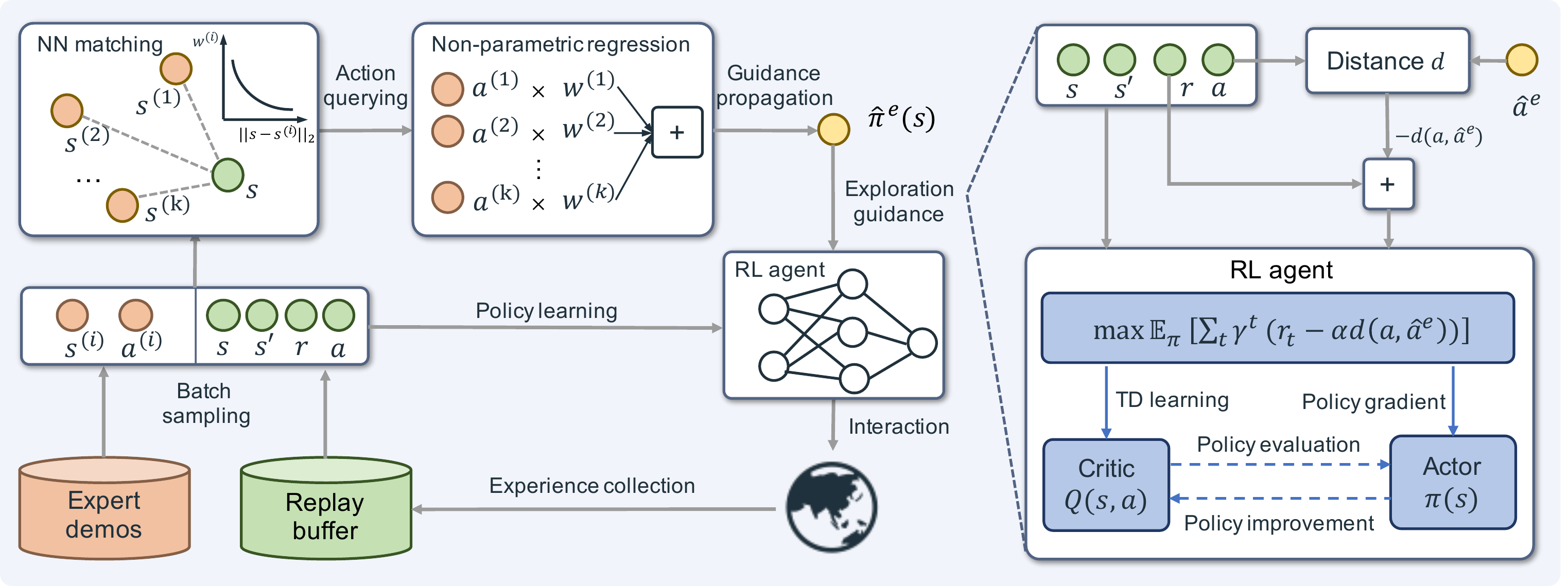}}
\caption{\textbf{RL with demonstrations-guided exploration.} We present the overall illustration of our proposed method DEX. It consists of two parts, a novel actor-critic based policy learning module that efficiently leverages demonstration date to guide the RL exploration (right), and a non-parametric module based on nearest-neighbor matching and locally weighted regression for robust guidance propagation at states far from demonstrated ones (upper left).}
\label{fig:framwork}
\vspace{-0cm}
\end{figure*}

We develop Demonstration-guided EXploration~(DEX), a novel exploration-efficient demonstration-guided RL algorithm for surgical subtask automation with limited demonstrations. Our method addresses the potential overestimation issue in existing methods based on our proposed actor-critic framework in Section~\ref{subsec:drrl}. To offer exploration guidance at states unobserved in demonstration data, our method propagates the guidance from demonstrations through a non-parametric module in Section~\ref{subsec:lwl}. Figure~\ref{fig:framwork} illustrates the overall framework of the proposed method DEX.

\noindent\textbf{Problem Formulation: }We consider an off-policy RL agent that interacts within an environment formulated by a Markov decision process. At time step $t$, the agent takes an action $a_t$ based on the current state $s_t$ and its deterministic policy $\pi$. The environment then rewards the agent with $r_t:=r(s_t,a_t)$ and transits it to the successor state $s_{t+1}$. After each transition, the agent stores the experience $(s_t, a_t, r_t, s_{t+1})$ into a replay buffer $\abuff$. Meanwhile, it also maintains a separate demonstration buffer $\ebuff$, where experiences are collected through an unavailable expert policy $\pi^e$ in advance.

\subsection{Expert-Guided Actor-Critic Framework}\label{subsec:drrl}
Many existing actor-critic based methods learn optimal policy by training an actor that maximizes the expected return $\mathbb{E}_\pi[\sum_{t=0}^\infty\gamma^tr_t]$, which is estimated by a critic that approximates the Q-value function, where $\gamma\in(0,1]$ is a discount factor. Their common way to guide exploration is to jointly maximize the Q-value and the behavioral similarity between the agent and expert. Effective as it is, the inaccurate Q-value estimates of expert-unlike actions may dominate the guidance embodied as behavioral dissimilarity, and thus hamper the exploration, reminiscent of the regularization issue in offline RL literature~\cite{kostrikov2021offline,fujimoto2019off}. In other words, the agent is likely to take unproductive actions due to value misestimate, which hinders it from exploring expert-like actions. 

To remedy this issue, we introduce a novel RL objective that augments the environment reward with the behavior gap between agent policy and expert policy:
\begin{align}\label{eq:obj}
    \max_{\pi} \mathbb{E}_\pi\left[\sum_{t=0}^{\infty} \gamma^t(r_t-\alpha d(a_t, \eaction_t)) \right], \;\; \eaction_t:=\epolicy(s_t),
\end{align}
where $\alpha$ is an exploration coefficient and the function $d(\cdot,\cdot)$ is a distance metric that measures the behavioral similarity between agent action and expert action. Correspondingly, we define the regularized Q-value function as $Q^\pi(s,a) := \mathbb{E}_\pi\left[\sum_{t=0}^{\infty} \gamma^t(r_t-\gamma\alpha d(a_{t+1}, \eaction_{t+1}))\big|s, a \right]\label{eq:qvalue}$. It assigns higher value estimates to actions similar to the expert action at any given state. It thus mitigates the effect of the inaccurate estimates of undesired actions by lowering its value through a dissimilarity regularization. Meanwhile, echoed with SAC~\cite{sac}, an appropriate choice of $\alpha$ steers the exploration-exploitation balance and benefits policy learning.

Based on the formulation, we introduce our proposed expert-guided actor-critic frameworks. The regularized value function (critic) $Q_\theta$, parameterized by $\theta$, minimizes the squared Bellman residual:
\begin{align}\label{eq:closs}
    \mathcal{L}_Q(\theta) = \mathbb{E}_{\abuff\cup\ebuff}\left[ \left(r_t+\gamma V_{\bar{\theta}}(s_{t+1}) - Q_\theta(s_t,a_t)\right)^2\right],
\end{align}
where $V_{\bar{\theta}}$ is state value function served as Q-target whose parameters $\bar{\theta}$ are the exponential moving average of $\theta$ that stabilize the training process. The actor, parametrized by $\phi$, is trained by maximizing following the objective:
\begin{align}\label{eq:aloss}
    \mathcal{L}_\pi(\phi) = \mathbb{E}_{\abuff\cup\ebuff}\left[Q_\theta(s_t, \pi_\phi(s_t))-\alpha d(\pi_\phi(s_t), a_t^e))\right].
\end{align}

Our framework bears some resemblances to behavior regularized actor-critic in offline RL domain~\cite{awac,brac}. At the online fine-tuning stage, their formulas target conservative policy updates by constraining the behavior gap between old and new policies. Different from their focus, our framework treats expert policy as a static reference policy and uses the behavior gap between agent and expert as a regularization term to accelerate the RL exploration more efficiently. 

\subsection{Guidance Propagation from Limited Demonstrations}\label{subsec:lwl}
Prior methods provide the actor with exploration guidance only at states observed in demonstration data, since the expert action $a_t^e$ is available only therein. However, the agent is prone to visit regions uncovered by demonstrations at the initial stage of training, where the demonstrations are incapable of supervising the actor exploration. On the other hand, such an issue also limits the feasibility and efficacy of critic regularization. Only regularizing the demonstration-covered Q-value introduces an underestimation bias on the value estimate to these state-action pairs due to the negativity of the regularization term. This negates the original purpose of enforcing critic regularization. Propagating guidance from limited demonstrations is thus of high necessity to efficiently realize the proposed actor-critic framework.

One natural solution is to learn a parametric expert policy approximator $\hat{\pi}^e$ from demonstrations in an parametric manner, e.g., behavior cloning. Although using a parametric model already outperforms existing methods, which demonstrates the effectiveness of our framework, we empirically observe that the parametric propagated guidance, i.e., $\hat{\pi}(s)$, fairly differs from the ground-truth expert action when states are far from the demonstration ones. We thus resort to a non-parametric regression model that empirically propagates more robust guidance from limited demonstrations. 

Specifically, we first sample a minibatch of states and actions from demonstration buffer $\ebuff$. Then, given any state $s$, we search states in the minibatch and find $k$ nearest neighbors of $s$ based on the Euclidean distance. The selected $k$ states and their associated actions are denoted by $\{(s^{(i)}, a^{(i)})\}^k_{i=1}$, a set of transitions from $\ebuff$. Subsequently, we approximate the expert policy through locally weighted regression~\cite{atkeson1997locally} with exponential kernel function: 
\begin{align}\label{eq:lwl}
    \hat{\pi}^e(s) = \frac{\sum_{i=1}^k \mathrm{exp}\left(-\|s-s^{(i)}\|_2\right)\cdot a^{(i)}}{\sum_{i=1}^k \mathrm{exp}\left(-\|s-s^{(i)}\|_2 )\right)}.
\end{align}
The assumption here is that similar states share similar optimal actions, whose effectiveness and robustness have been demonstrated in some robotic learning tasks~\cite{vinn,fist,dime}. Unlike these IL methods, we aim to use the limited demonstration data to accelerate the exploration of RL and increase the generalization ability of policy.  

\subsection{Implementation Details} 
We implement our method with deep deterministic policy gradient (DDPG,~\cite{ddpg}) from OpenAI Baselines~\cite{baselines} and adopt their main hyperparamter settings. We choose DDPG to avoid the potential pathology of density-based methods~\cite{rudner2021pathologies}, though our implementation can be naturally extended to the stochastic setting.
Specifically, both actor and critic are parameterized as four fully-connected layers of hidden dimension 256 interpolated with \texttt{ReLU} activations, where actions are scaled to the range [-1,1] by a \texttt{Tanh} activation in the actor network. They are trained with ADAM~\cite{adam}. The exploration noise is set as Gaussian noise with a scale factor of 0.1,  For goal-conditioned environments, we adopt hindsight experience replay~\cite{her} with future sampling strategy on 80\% examples for both replay buffer and demonstration buffer. The distance metric is set as $L2$-norm for all tasks. The coefficient of exploration term $\alpha$ is set as $5$ for all tasks. The number of nearest neighbors $k$ used for expert policy estimation is set to $5$. More analyses on $\alpha$ and $k$ would be given in the ablation studies.

\section{Experiments}
In this section, we first conduct experiments on SurRoL~\cite{surrol}, a comprehensive simulation environment for task automation of surgical robots. Extensive evaluation and ablation analysis on 10 challenging surgical manipulation tasks demonstrate the effectiveness of our method DEX for surgical robot learning. Moreover, we transfer the learned policies to a dVRK platform for real-world robotic deployment. 

\begin{figure}[t]
\centering
\centerline{\includegraphics[width=1\linewidth]{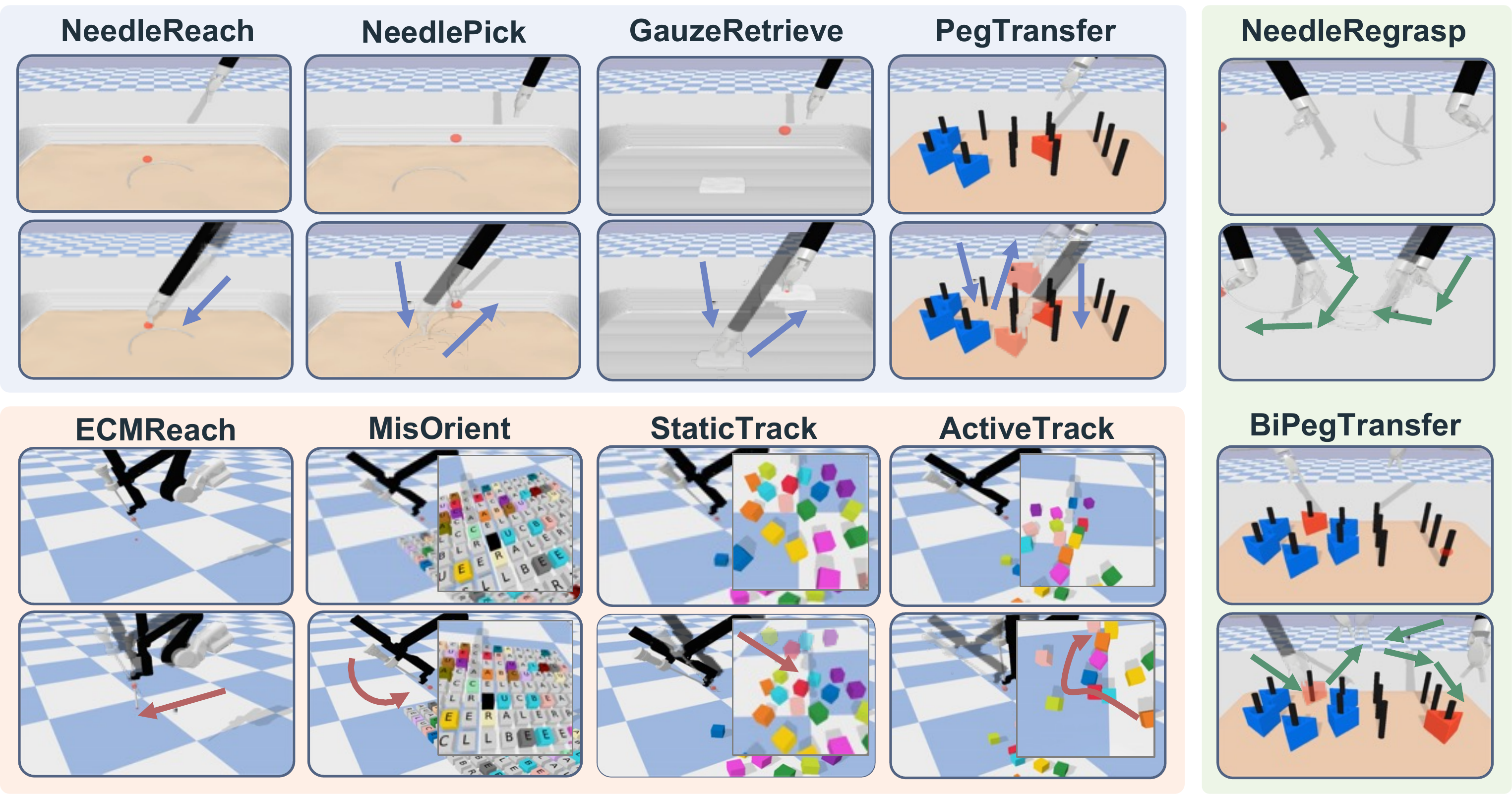}}
\caption{\textbf{Task description of SurRoL.} We choose ten surgical tasks from SurRoL and divide them into three domains according to the type of manipulators. The arrows in each task represent the task flow as well as the execution process of the scripted expert policy provided by the platform.}
\label{fig:surrol}
\end{figure}

\begin{table*}
\centering
 \caption{\textbf{Performance after training 100k environment steps}. We present mean scores with standard deviations of each task and aggregate IQMs with 95\% stratified bootstrap CIs of each domain (grey-colored cells). Results are over 10 runs with different seeds, where each run averages 20 evaluations. We abbreviate sparse and dense reward to $S$ and $D$, respectively. Our method achieves prominent manipulation performance on complex tasks and domains.}
  \label{tab:main}
  \vspace{-0.6mm}
 \resizebox{1\textwidth}{!}
 { 
  \begin{tabular}{clccccccccccc}

    \multicolumn{3}{c}{Task Description} & \multicolumn{2}{c}{Reinforcement Learning} & \multicolumn{3}{c}{Imitation Learning} & \multicolumn{4}{c}{Demonstration-guided Reinforcement Learning} & Ours\\[0.5ex]

    \toprule
    & Task  & $\mathcal{S}/\mathcal{A}/r$ & SAC~\cite{Haarnoja2018SoftAA}    & DDPG~\cite{ddpg}  & BC~\cite{bc} & SQIL~\cite{sqil}  & VINN~\cite{vinn}  & DDPGBC~\cite{ddpgher} & AMP~\cite{amp} & CoL~\cite{col} & AWAC~\cite{awac} & DEX\\

    \midrule
     \parbox[c]{1.2mm}{\multirow{5}{*}{\rotatebox[origin=c]{90}{ECM}}} 
    &  \ourcell Aggregate  & \ourcell -- & \textbf{\ourcell 0.99\ci{.03}}  & \ourcell  \textbf{0.99\ci{.02}}  &  \ourcell \textbf{1.00\ci{.00}} & \ourcell 0.24\ci{0.06}  & \ourcell 0.58\ci{.06}  & \ourcell \textbf{1.00\ci{.00}}  & \ourcell \textbf{1.00\ci{.01}}  & \ourcell \textbf{1.00\ci{.00}}  & \ourcell \textbf{0.99\ci{.01}}  & \ourcell \textbf{1.00\ci{.00}}\\[0.5ex]
     
    \cdashline{2-13}\noalign{\vskip 0.5ex}
    & ECMReach & $\mathbb{R}^{12}/\mathbb{R}^{3}/S$ & \textbf{1.00\ci{.06}}  & \textbf{1.00\ci{.00}} & \textbf{1.00\ci{.00}} & 0.07\ci{.04}  & 0.49\ci{.10}  & \textbf{1.00\ci{.00}} & \textbf{0.99\ci{.02}} & \textbf{1.00\ci{.00}} & \textbf{1.00\ci{.00}} & \textbf{1.00\ci{.00}}\\
    & StaticTrack & $\mathbb{R}^{16}/\mathbb{R}^{3}/S$ & 0.92\ci{.14}  & \textbf{0.98\ci{.05}} & \textbf{1.00\ci{.00}} & 0.43\ci{.26}  & 0.56\ci{.10}  & \textbf{1.00\ci{.00}} & \textbf{0.97\ci{.03}} & \textbf{1.00\ci{.00}} & \textbf{1.00\ci{.00}} & \textbf{1.00\ci{.00}}\\
    & MisOrient & $\mathbb{R}^{11}/\mathbb{R}^{1}/S$ & \textbf{1.00\ci{.00}}  & \textbf{1.00\ci{.00}}  & \textbf{1.00\ci{.00}} & 0.56\ci{.10}  & 0.50\ci{.11}  & \textbf{0.99\ci{.02}} & \textbf{0.98\ci{.02}} & \textbf{0.99\ci{.02}} & \textbf{0.98\ci{.03}} & \textbf{0.99\ci{.02}}\\
    & ActiveTrack & $\mathbb{R}^{10}/\mathbb{R}^{3}/D$ &   0.79\ci{.08}  & 0.67\ci{.08}  & \textbf{0.95\ci{.01}} & 0.07\ci{.06}  & \textbf{0.92\ci{.06}}  & 0.81\ci{.05} & \textbf{0.94\ci{.01}} & \textbf{0.96\ci{.01}} & 0.51\ci{.12} & \textbf{0.94\ci{.01}}\\

    \cmidrule(lr){2-13}
    \parbox[c]{1.2mm}{\multirow{5}{*}{\rotatebox[origin=c]{90}{PSM}}} 
    &  \ourcell Aggregate & \ourcell -- & \ourcell 0.0\ci{.00}  & \ourcell  0.00\ci{.00}  &  \ourcell 0.40\ci{.05} & \ourcell 0.00\ci{.00}  & \ourcell 0.02\ci{.02}  & \ourcell 0.80\ci{.04} & \ourcell 0.00\ci{.00}  &  \ourcell 0.85\ci{.06}  &  \ourcell 0.46\ci{.19}  & \ourcell \textbf{0.89\ci{.03}}\\[0.5ex]
     
    \cdashline{2-13}\noalign{\vskip 0.5ex}
    & NeedleReach & $\mathbb{R}^{13}/\mathbb{R}^5/S$ & \textbf{1.00\ci{.00}}  & \textbf{1.00\ci{.00}}  & \textbf{1.00\ci{.00}} & 0.07\ci{.09}  & 0.89\ci{.06}  & \textbf{1.00\ci{.00}} & \textbf{0.99\ci{.02}} & \textbf{1.00\ci{.00}} & 0.94\ci{.20} & \textbf{1.00\ci{.00}} \\
    & GauzeRetrieve & $\mathbb{R}^{25}/\mathbb{R}^5/S$ & 0.00\ci{.00}  & 0.00\ci{.00}  & 0.07\ci{.05} & 0.00\ci{.00}  & 0.01\ci{.02}  & 0.63\ci{.11} & 0.00\ci{.00} & \textbf{0.71\ci{.16}} & 0.43\ci{.43} & \textbf{0.73\ci{.12}} \\
    & NeedlePick & $\mathbb{R}^{25}/\mathbb{R}^5/S$ & 0.00\ci{.00}  & 0.00\ci{.00}  & 0.21\ci{.06} & 0.00\ci{.00}  & 0.02\ci{.02}  & \textbf{0.91\ci{.05}} & 0.00\ci{.00} & \textbf{0.96\ci{.05}} & 0.26\ci{.33} & \textbf{0.94\ci{.05}} \\
    & PegTransfer & $\mathbb{R}^{25}/\mathbb{R}^5/S$ & 0.00\ci{.00}  & 0.00\ci{.00}  & 0.56\ci{.11} & 0.02\ci{.05}  & 0.05\ci{.04}  & 0.48\ci{.22} & 0.00\ci{.00} & 0.58\ci{.23} & 0.31\ci{.32} & \textbf{0.73\ci{.20}}\\

    \cmidrule(lr){2-13}
     \parbox[c]{1.2mm}{\multirow{3}{*}{\rotatebox[origin=c]{90}{Bi-PSM}}} 
     & \ourcell Aggregate & \ourcell -- & \ourcell 0.00\ci{.00}  & \ourcell 0.00\ci{.00}  & \ourcell 0.08\ci{.04} & \ourcell 0.00\ci{.00}  &  \ourcell 0.00\ci{.00}  & \ourcell 0.00\ci{.00} &  \ourcell 0.00\ci{.00}  &  \ourcell 0.00\ci{.00}  &  \ourcell 0.00\ci{.00}  & \ourcell \textbf{0.39\ci{.11}}\\[0.5ex]
     
    \cdashline{2-13}\noalign{\vskip 0.5ex}
    & NeedleRegrasp & $\mathbb{R}^{41}/\mathbb{R}^{10}/S$ & 0.00\ci{.00}  & 0.00\ci{.00}  & 0.09\ci{.03}  & 0.01\ci{.00}  & 0.01\ci{.02}  &  0.05\ci{.08} & 0.00\ci{.00} & 0.04\ci{.07} & 0.00\ci{.00} & \textbf{0.63\ci{.19}} \\
    & BiPegTransfer & $\mathbb{R}^{41}/\mathbb{R}^{10}/S$ & 0.00\ci{.00}  & 0.00\ci{.00}  & 0.09\ci{.05} & 0.00\ci{.00}  & 0.00\ci{.00}  & 0.00\ci{.00} & 0.00\ci{.00} & 0.01\ci{.02} & 0.00\ci{.00} & \textbf{0.18\ci{.14}}\\
    
    \midrule
    &  \ourcell Overall & \ourcell -- & \ourcell 0.46\ci{.03}  & \ourcell  0.45\ci{.01}  &  \ourcell 0.68\ci{.02} & \ourcell 0.02\ci{.02}  & \ourcell 0.24\ci{.03}  & \ourcell 0.83\ci{.05} &   \ourcell 0.48\ci{.01}  &  \ourcell 0.87\ci{.03}  &  \ourcell 0.58\ci{.08}  &  \textbf{\ourcell 0.92\ci{.02}}\\
    
    \bottomrule
    \vspace{-0.7cm}
  \end{tabular}

  }
\end{table*}

\subsection{Experimental Setup}\label{subsec:setup}

\noindent\textbf{Different Tasks on Surgical Manipulation.} To evaluate whether our method DEX can effectively handle different surgical manipulation scenarios, we conduct experiments on 10 challenging tasks from SurRoL. These 10 tasks widely cover three different domains according to the type of manipulator: (1) Single-handed patient-sided manipulator (\textbf{PSM}); (2) Bimanual PSM (\textbf{Bi-PSM}); and (3) Endoscopic camera manipulator (\textbf{ECM}). Figure~\ref{fig:surrol} illustrates the involved 10 manipulation tasks. All tasks are goal-conditioned with sparse reward functions that indicate task success, except for \emph{ActiveTrack}, whose reward function is dense that depends on the precision of object tracking. We use low-dimensional state representation that consists of object state (3D Cartesian positions and 6D pose) and robot proprioceptive state (jaw status and end-effector position), and Cartesian-space control as action space. Please refer to~\cite{surrol} for more details on the different surgical manipulation tasks. Besides, we sample 100 successful episodes of demonstration data for each task through the scripted expert policy provided by SurRoL.

\noindent\textbf{Evaluation Metrics.} Similar to~\cite{curl,drq}, we measure the manipulation performance of each task after a uniform 100k environment steps. The manipulation score for each task is linearly re-scaled to be within $[0,1]$ for comparisons across tasks as well as better visualization. Following~\cite{rliable}, we adopt interquartile mean~(IQM) and 95\% interval estimates via stratified bootstrap confidence intervals~(CIs) to measure the aggregate performance over each domain robustly.

\noindent\textbf{Baselines.} First of all, we compare our method with two state-of-the-art pure RL algorithms without using demonstrations: (1) \textbf{SAC}~\cite{sac} and (2) \textbf{DDPG}~\cite{ddpg}. Besides, we compare our method with three representative IL methods: (3) \textbf{BC}~\cite{bc} that supervises the agent to imitate expert actions from demonstrations. (4) \textbf{SQIL}~\cite{sqil} that reduces IL to RL by assigning positive rewards to demonstrations, a representative of the state-of-the-art IL algorithms based on SAC. (5) \textbf{VINN}~\cite{vinn} that directly uses the estimate of expert policy in Equation~\eqref{eq:lwl} as the agent policy. Moreover, we compare our method with four existing demonstration-guided RL algorithms: (6) \textbf{DDPGBC}~\cite{ddpgher} that guides DDPG actor with Q-filtered BC loss. (7) \textbf{AMP}~\cite{amp} that augments environment with GAIL~\cite{gail} reward based on SAC. (8) \textbf{CoL}~\cite{col} that initializes the policy with BC offline and integrates BC with DDPG in the online training stage. (9) \textbf{AWAC}~\cite{awac} that trains an offline RL with demonstrations and online fine-tunes it with a conservative constraint between current and old policies. 

\begin{table}
\vspace{6mm}
\centering
    \caption{\textbf{Ablation on demonstration amount.} We offer our method and baselines different amounts of demonstrations. Results are IQMs with 95\% stratified bootstrap CIs of 5 runs over PSM and Bi-PSM domains.}
    \label{tab:main_demo}
    \vspace{-0.2mm}
    \setlength{\tabcolsep}{5pt}
    \renewcommand{\arraystretch}{1.05}
    \resizebox{1\linewidth}{!}
    { 
    \begin{tabular}{lccccc}
    \toprule\noalign{\vskip -0.2ex}
    \multirow{2}{*}{Method}  & \multicolumn{5}{c}{Number of episodes in demonstrations}  \\ \noalign{\vskip -0.7ex}\cmidrule(lr){2-6}\noalign{\vskip -0.7ex}
    & 10 epi.  & 25 epi.  & 50 epi.  & 75 epi. & 100 epi.\\ 
        
    \noalign{\vskip -0.5ex}\midrule
    BC~\cite{bc} & 0.00\ci{.00} & 0.05\ci{.02} & 0.15\ci{.03} & 0.19\ci{.03} & 0.22\ci{.03} \\
    SQIL~\cite{sqil} & 0.00\ci{.00} & 0.00\ci{.00} & 0.00\ci{.00} & 0.00\ci{.00} & 0.00\ci{.00} \\
    VINN~\cite{vinn} & 0.03\ci{.03} & 0.02\ci{.03} & 0.02\ci{.02} & 0.01\ci{.02} & 0.01\ci{.01} \\
    \cdashline{1-6}\noalign{\vskip 0.5ex}
    DDPGBC~\cite{ddpgher} & 0.00\ci{.03} & 0.20\ci{.11} & 0.35\ci{.05} & 0.47\ci{.10} & 0.45\ci{.08} \\
    AMP~\cite{amp} & 0.00\ci{.00} & 0.00\ci{.00} & 0.00\ci{.00} & 0.00\ci{.00} & 0.00\ci{.00} \\
    CoL~\cite{col} & 0.18\ci{.07} & 0.47\ci{.09} & 0.46\ci{.09} & 0.49\ci{.05} & 0.43\ci{.10} \\
    AWAC~\cite{awac} & 0.00\ci{.05} & 0.05\ci{.14} & 0.07\ci{.18} & 0.40\ci{.15} & 0.39\ci{.13} \\
    \noalign{\vskip -0.2ex}\midrule\noalign{\vskip -0.2ex}
    \ourcell DEX (ours) & \ourcell \textbf{0.24}\ci{.02} & \ourcell \textbf{0.50}\ci{.09} & \ourcell \textbf{0.68}\ci{.07} & \ourcell \textbf{0.78}\ci{.07} & \ourcell \textbf{0.80}\ci{.06} \\
    \noalign{\vskip -0.2ex}
    \bottomrule
    
    \vspace{-0.8cm}
    \end{tabular}
    }
\end{table}

\subsection{Main Results on SurRoL Platform}\label{subsec:main_results}
We present the aggregate and distributed performance of each method in Table~\ref{tab:main}. The results show that two pure RL methods achieve great performance in ECM domain, but poor performance in PSM and Bi-PSM. This implies that their exploration strategies are not applicable in complex surgical manipulation tasks, while our demonstration-depend exploration is able to handle these tasks. Similar to pure RL, three IL methods perform well in ECM domain, while their performance decreases dramatically in the PSM and Bi-PSM domain, which has a higher dimension of state and action space and requires more complex manipulation skills. This indicates that, given a limited number of demonstrations, the generalization ability of IL is severely deteriorated by the complexity of the task. Compared to IL, our method not only perform similarly great in ECM domain, but achieves \textbf{2.2x} IQM in PSM domain and, especially, \textbf{4.9x} in Bi-PSM domain. This demonstrates the advances of our method that leverages demonstrations to accelerate RL exploration.

Meanwhile, we also observe that three demonstration-guided RL methods, except for AMP which suffers from the instability issue of adversarial training~\cite{osa2018algorithmic}, perform favorably in ECM and PSM domains. This may attribute to their utilization of demonstration for RL exploration. However, these methods still get stuck in the complex PSM tasks, such as \emph{PegTransfer}, let alone the Bi-PSM domain that additionally demands coordination skills between two manipulators. Compared with them, our method fully leverages demonstrations to mitigate the guidance impediment of critic and guide the exploration efficiently at demonstration-uncovered states, leading to a prominent performance improvement of \textbf{0.39} in the Bi-PSM domain.

\begin{figure}[t]
\centering
\vspace{1mm}
\centerline{\includegraphics[width=1\linewidth]{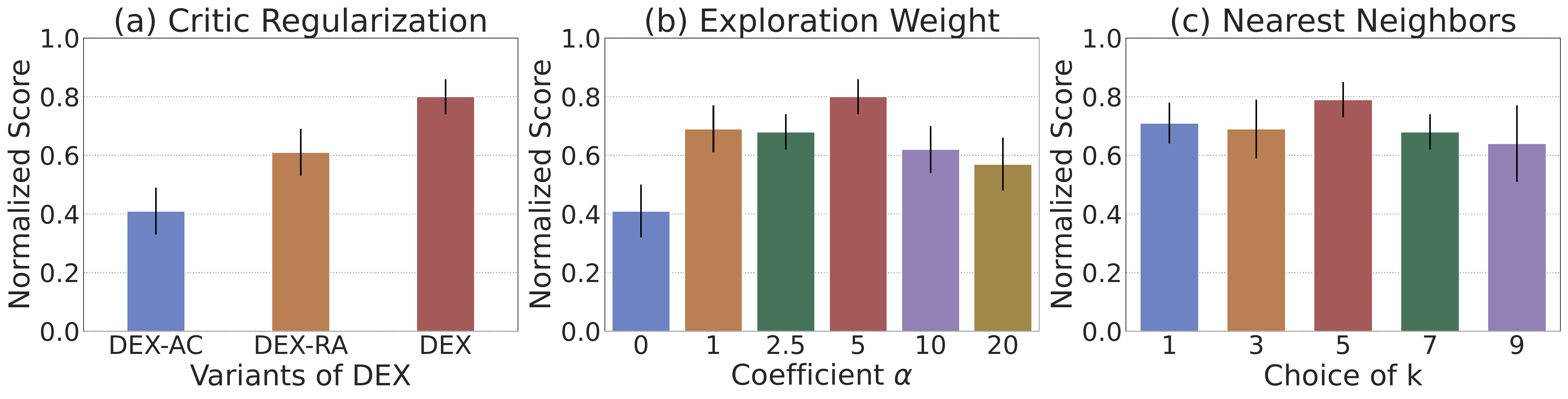}}\vspace{0mm}
\caption{\textbf{Ablation studies.} In (a), we propose two variants of \ourmethod~that partially regularize the actor-critic to investigate the effectiveness of our proposed critic regularization. In (b), we ablate the choice of exploration coefficient $\alpha$. In (c), we test the influence of the number of nearest neighbors $k$ in non-parametric propagation. All results are IQMs with 95\% stratified bootstrap CIs (error bar) over 5 runs on PSM and Bi-PSM domains. }
\label{fig:ablations}
\vspace{-0.2mm}
\end{figure}

\begin{figure*}[t]
\centering
\centerline{\includegraphics[width=1\textwidth]{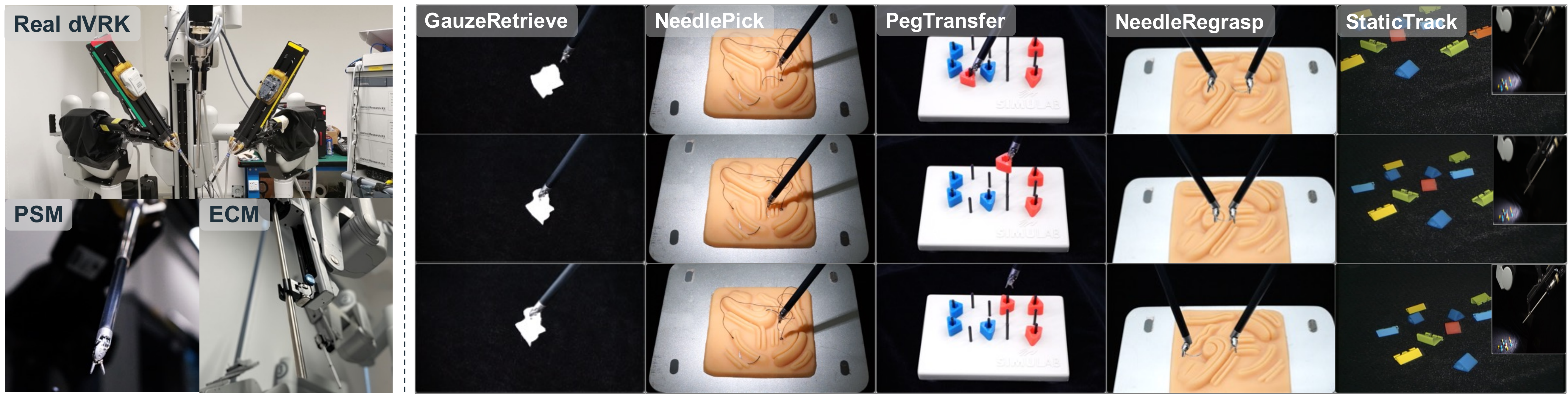}}
\caption{\textbf{Deployment on real dVRK.} We deploy the learned policies on the real dVRK platform (left) and present the generated trajectories of five representative tasks (right). The successful deployment demonstrates the transfer ability of our method from simulation to real-world surgical scenarios.}
\vspace{0cm}
\label{fig:dvrk}
\end{figure*}

\subsection{Ablation Studies}\label{subsec:abl}
\noindent\textbf{Effect of Demonstration Amount.} We further study the manipulation performance of each method with different amounts of demonstrations. We train agents with $10\%$, $25\%$, $50\%$, $75\%$ and $100\%$ of demonstrations, respectively. Table~\ref{tab:main_demo} presents the aggregate results over PSM and Bi-PSM domains. Our proposed method DEX consistently outperforms other competing methods with different amounts of demonstrations. This result indicates that with the same amount of demonstrations, our DEX can utilize demonstrations more efficiently to achieve a higher manipulation performance. More importantly, our proposed DEX exhibits a lower demand for the number of demonstrations for task automation. Even with merely $25\%$ of demonstrations, DEX has outperformed most competing methods that are trained with $100\%$ of demonstrations. This characteristic demonstrates that DEX is well suited for the challenging surgical scenarios where collecting large amounts of demonstrations is unaffordable.

\noindent\textbf{Effect of Critic Regularization.} While some existing methods only impose the guidance on actor~\cite{ddpgher,col,chiu2021bimanual}, our method guide both actor and critic to remedy the guidance impediments of the ordinary critic. In order to further investigate the effects of critic regularization on the manipulation performance, we propose two variants of \ourmethod~that partially regularize the actor-critic modules in Figure~\ref{fig:ablations}(a). The results show that, compared with the ordinary no-guidance actor-critic (DEX-AC), the variant that only regularizes actor (DEX-RA) brings an improvement of 0.20 w.r.t. IQM, 
and the proposed one which regularizes both actor and critic (DEX) brings 0.39. It demonstrates the impediment of inaccurate value estimates of ordinary critic and the effectiveness of our proposed critic regularization to alleviate this issue.

\noindent\textbf{Effect of Exploration Coefficient $\alpha$.} The exploration coefficient $\alpha$ determines the relative importance of the exploration term against environment reward. We investigate the effect of this parameter by gradually increasing the value of $\alpha$ from $0$ to $20$ and repeatedly train the agent with the remaining settings identical. Figure~\ref{fig:ablations}(b) present the experiment results. It shows that $\alpha$ around $5$ yields the best performance, indicating that an intermediate choice balances exploration and exploitation well. We also note that an adaptive coefficient (e.g., decay $\alpha$) may further balance the trade-off to avoid local minima and leave it as future work.

\noindent\textbf{Effect of Guidance Propagation.} To investigate the effect of guidance propagation, we propose a parametric variant (DEX-BC) that pre-trains a BC model to approximate the expert action at demonstration-uncovered states. The results show that, compared with no-propagation methods, DEX-BC still achieves a performance improvement of 0.3 IQM over PSM and Bi-PSM domains, though is 0.05 lower than DEX. It demonstrates not only the robustness of non-parametric guidance propagation but the insensitivity to the choice of propagation method in our actor-critic framework. 

\noindent\textbf{Effect of $k$ for Guidance Propagation.} We investigate the influence of the number of nearest neighbors $k$ by gradually increasing its value from 1 to 9. The results in Figure~\ref{fig:ablations}(c) show that too large value of $k$ leads to a slight performance drop and an intermediate value around 5 works well, indicating that our method is not sensitive to the choice of $k$ within an appropriate interval. 

\begingroup
\setlength{\tabcolsep}{3pt}
\begin{table}[t]
\vspace{7mm}
\caption{\textbf{Evaluation on real dVRK.} We test success rate of the learned policies with 20 trials for each task and compare with two baselines.}
\label{tab:dvrk}
\centering
\resizebox{0.5\textwidth}{!}{%
\begin{tabular}{lcccccc}
\toprule
\textbf{Method}  & GauzeRetrieve & NeedlePick & PegTransfer &  NeedleRegrasp & StaticTrack  \\\midrule
BC~\cite{bc}           & $0.00$ & $0.85$   & $0.00$ & $0.40$ & $1.00$ \\
DDPGBC~\cite{ddpgher}      & $0.75$ &$0.95$  & $0.35$ & $0.65$ & $1.00$ \\
\midrule
\ourmethod~(ours)   & $\mathbf{0.90}$ & $\mathbf{0.95}$  & $\mathbf{0.75}$ & $\mathbf{0.90}$  & $\mathbf{1.00}$ \\
\bottomrule
\end{tabular}}
\vspace{-0.6cm}
\end{table}
\endgroup

\subsection{Deployment on Real Robot Platform of dVRK}\label{subsec:real_dvrk}
To validate the transfer ability of our method, we train RL policies in SurRoL and deploy them on dVRK systems. We conduct real-world manipulation experiments on five representative tasks, namely \emph{GauzeRetrieve}, \emph{NeedlePick} and \emph{PegTransfer} in PSM domain, \emph{NeedleRegrasp} in Bi-PSM domain, and \emph{StaticTrack} in ECM domain. These tasks are selected to be diverse and comprehensive, requiring different levels of surgical manipulation skills. We train policies with 300k steps within SurRoL and select checkpoints with the best performance for deployment. We present experiment snapshots in Figure~\ref{fig:dvrk}. In three PSM tasks, we observe that agents successfully learn skills of reaching, picking, and placing to accomplish the task without collisions. In \emph{NeedleRegrasp}, the policy additionally acquires coordination skills between two manipulators compared with policies in the PSM domain. In \emph{StaticTrack}, the trained policy successfully learns endoscopic control with a high success rate. Meanwhile, we also compare our method with one imitation learning method BC and one competitive demonstration-guided RL method DDPGBC. The results in Table~\ref{tab:dvrk} show that our method achieves the highest success rates at all tasks with 20 trials, which demonstrates the great potential of our method for real-world surgical task automation.

\section{Conclusion}
\label{sec:conclusion}
We present DEX, a novel demonstration-guided RL method that narrows down exploration space by encouraging expert-like behaviors and enabling robust guidance when confronting states unobserved in demonstrations, to improve exploration efficiency with a modest set of demonstrations. We first demonstrate its performance improvement on 10 challenging surgical manipulation tasks compared with state-of-the-art methods on the SurRoL platform. We also deploy the trained policy to the dVRK platform to show its potential for transferring to real-world surgical automation scenarios.

{\footnotesize
\bibliographystyle{IEEEtranN}
\bibliography{ref}
}

\end{document}